\documentclass{article}
\usepackage{spconf,amsmath,graphicx,multirow,color,booktabs}

\title{LiteG2P: A fast, light and high accuracy model for Grapheme-to-Phoneme conversion}
\name{Chunfeng Wang, Peisong Huang, Yuxiang Zou, Haoyu Zhang, Shichao Liu, Xiang Yin, Zejun Ma}
\address{Bytedance AI-Lab, China}

\begin{document}
%
\maketitle
\begin{abstract}
As a key component of automated speech recognition (ASR) and the front-end in text-to-speech (TTS), grapheme-to-phoneme (G2P) plays the role of converting letters to their corresponding pronunciations. Existing methods are either slow or poor in performance, and are limited in application scenarios, particularly in the process of on-device inference. In this paper, we integrate the advantages of both expert knowledge and connectionist temporal classification (CTC) based neural network and propose a novel method named LiteG2P which is fast, light and theoretically parallel. With the carefully leading design, LiteG2P can be applied both on cloud and on device. Experimental results on the CMU dataset show that the performance of the proposed method is superior to the state-of-the-art CTC based method with 10 times fewer parameters, and even comparable to the state-of-the-art Transformer-based sequence-to-sequence model with less parameters and 33 times less computation.
\end{abstract}
\section{Introduction}
\label{sec:intro}

Grapheme-to-phoneme(G2P) aims to convert word letters to its pronunciations, and is widely used in speech tasks, such as speech synthesis \cite{wang17n_interspeech,ren2019fastspeech,zou2021fine} and speech recognition \cite{amodei2016deep,dong2020cif}. With the rapid development of speech application scenarios, especially for TTS, there is a trend of higher requirements for both speed and performance in the G2P task.

There exists decades of development for the research on the G2P task. In traditional series of methods, \cite{black1998issues} proposed the most early rule based model, while latter methods\cite{bisani2008joint,galescu2002pronunciation} were proposed to learn a joint ngram model as well as weighted finite state transducer (WFST) \cite{novak2012improving,novak2013failure}.

Recent years, more researchers focus on the end-to-end solutions with the development of deep neural models. The key problem of the G2P task is to find the alignment between word letters and phonemes. Based on this insight, many researchers regard the G2P task as a neural machine translation task so that existing sequence-to-sequence models are able to be utilized,  \cite{yolchuyeva2019grapheme} proposed series of encoder-decoder with attention models whose encoder and decoder are composed of convolutional layers or BiLSTM layers. A transformer based sequence to sequence model is introduced into G2P in\cite{Yolchuyeva2019}, which shows better accuracy and model size balance. However, some works try to solve the problem from another point of view, \cite{rao2015grapheme} firstly proposed a connectionist temporal classification (CTC)\cite{graves2006connectionist} based method in the G2P task. CTC is widely used in the fields of optical character recognition (OCR)\cite{shi2016end} and speech recognition\cite{amodei2016deep}, which needs to ensure the output length is less than or equal to the input length. For G2P task, \cite{rao2015grapheme} proposed a delay strategy to make the length of input and output to be equal.

Traditional methods\cite{black1998issues,bisani2008joint,galescu2002pronunciation,novak2012improving,novak2013failure} suffer poor performance and require heavy expert experience. Attention based sequence-to-sequence\cite{yolchuyeva2019grapheme,Yolchuyeva2019} models are usually autoregressive in decoding stage, which limit the computation efficiency and have the risk of stop token collapse. Existing CTC based methods\cite{rao2015grapheme} require the output length to conduct the delay strategy and the model is too heavy, which restrict the application scenarios. Based on above analysis, in our opinion, a fast, compact and high accuracy G2P model is ready to come out. And according to our knowledge, we are the first to propose a G2P model which can be applied both on cloud and on device with state-of-the-art performance.

We call the proposed method LiteG2P, there are three contributions of LiteG2P:

1. LiteG2P is the first end-to-end G2P model intergrating little expert knowledge into data driven model using CTC loss function to find the alignment.

2. LiteG2P is parallel, light and over 30x faster than state-of-the-art methods, which makes it easy to be deployed both on cloud and on device.

3. Performance of LiteG2P is comparable to the state-of-the-art transformer-based sequence-to-sequence model.


\section{LiteG2P}
\label{sec:format}

\begin{figure}[h]
\centering
\includegraphics[scale=0.4]{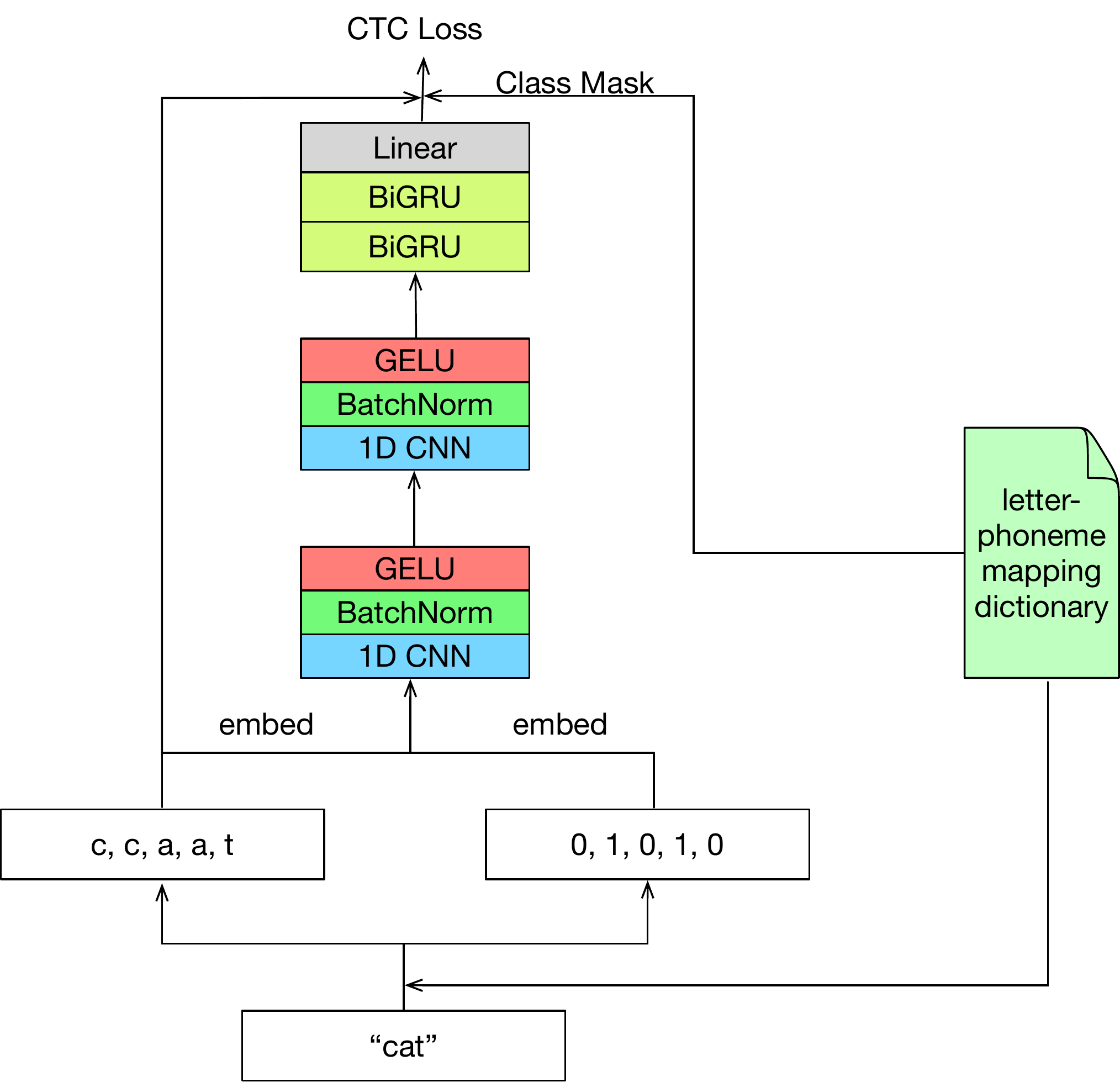}
\caption{The architechture of LiteG2P}
\label{fig1}
\end{figure}

Although sequence-to-sequence based models have shown good performance for the G2P task, we think they are redundant in that context. Being different from the neural machine translation task, G2P owns many inherent and inspiring features, such as monotonicity, limited mapping sets between letters and phonemes. And we find that the performance of sequence-to-sequence model is sensitive to the size in our experiments. Inspired by all above, we decide to design a new end-to-end architecture for the G2P task. G2P is a sequence transduction task in which an input orthographic sequence $W = [g1, g2, . . . , gn]$ is transformed to an output phoneme sequence $P = [p1, p2, . . . , pt]$ through a mapping function $f$, which is a neural network in this study. The whole architecture of LiteG2P is shown as Figure \ref{fig1}.
\vspace*{-10pt}

Without the help of attention mechanism in sequence-to-sequence model, we try to model the alignment between word letters and phonemes by using CTC \cite{graves2006connectionist} loss function, which interprets the
network outputs as a probability distribution over all possible
output label sequences, conditioned on the input data. The
CTC objective function directly maximizes the probabilities
of the correct labelings. Furthermore, we referenced a simple additional dictionary as expert knowledge which is shown in Table \ref{tab1}. The columns from left to right separately represent word letters, the corresponding maximum phoneme mapping length and possible mapping phonemes of US english. Different from \cite{rao2015grapheme}, in order to ensure the output length is less than or equal to the input length, we expand each letter to the corresponding max phoneme mapping length in data preprocessing. Moreover, we have found that each letter maps a limited phoneme set, which could reduce the search space of the model and make the model converge better. Thus, we utilize the possible mapping phonemes information as a mask in the output of the model to help the model learn.


\begin{table}
\caption{Letter and phoneme mapping dictionary of english}
\label{tab1}
\centering
\begin{tabular}{c|cc}
\toprule
Letter & Max mapping length & Possible phonemes \\
\midrule
a & 2 & AA0,AA1,AA2, ... \\
b & 1 & B,P \\
c & 2 & K,CH,S,SH,T \\
d & 1 & D,T,JH \\
e & 2 & IH0,IH1,IH2,... \\
... & ... & ... \\
x & 2 & K,S,G,Z,... \\
y & 1 & IY0,IY1,IY2,... \\
z & 2 & Z,T,S,... \\
\bottomrule
\end{tabular}
\end{table}

Before feeding the word letters into the model, the word is processed to get two parts of features as shown in Figure \ref{fig1}: the first part is the expansion letters whose expansion length is obtained from the dictionary. While in order to distinguish the repeated letters, local position embbedding is also considered, and in our experiments, we have found it brings slight performance improvement. After processing the embeddings, the two parts of features are concatenated together as the final input features. Specifically, the local position embedding is normalized as follows,

\vspace*{-20pt}
\begin{align}
pos_{j_{norm}}^{i} = \frac{j - len_{i}}{max(len_{i} - 1, 1)}
\end{align}

where $i$, $j$, $len_{i}$ denote the original letter index, the expansion local letter index which starts from one and the expansion length of the original $i$th letter.

\begin{figure*}[h]
\centering
\includegraphics[scale=0.75]{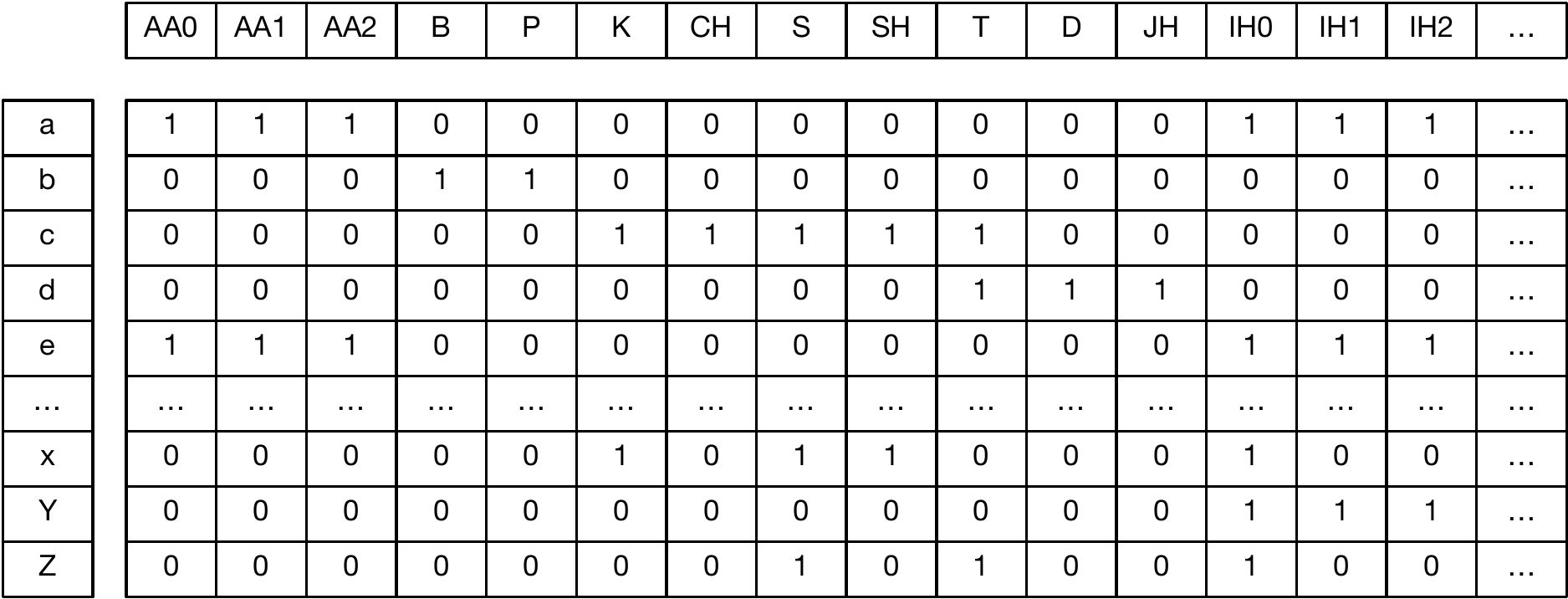}
\caption{The mask embedding matrix derived from the expert dictionary}
\label{fig2}
\end{figure*}

By referring to the most popular CTC based models in OCR \cite{shi2016end} and ASR \cite{amodei2016deep}, we design the model architecture of LiteG2P with 1D convolutional neural network (CNN) and bidirectional gated recurrent unit (GRU) \cite{cho-etal-2014-properties} layers. The model includes two CNN blocks, each block is composed of 1D CNN layer, batch normalization layer \cite{ioffe2015batch} and gaussian error linear units (GELU) \cite{hendrycks2016gaussian} activation function. And two bidirectional GRU layers follow the two CNN blocks. Finally, a mask embedding matrix as shown in Figure \ref{fig2} is derived from the expert dictionary and is fed into the final output of the model to help filtering unreasonable phoneme pairs. Each entry of the matrix is filled with 0 or 1, which represents whether the letter and the phoneme is related. The pseudo code of mask process is written as follows,

\vspace*{-20pt}
\begin{align}
   & output\_mask = embedding\_lookup(input, matrix_{mask}) \\
   & output = f(input,local\_pos) * output\_mask \\
   & output[output == 0] = INT\_MIN \\ 
   & output = softmax(output)
\end{align}
\vspace*{-20pt}

where $input$, $local\_pos$, $matrix\_mask$, $f$ denote expansion letters, local position embedding, the mask embedding matrix and model function, respectively. And after this mask process, for each letter, the probability of irrelevant pronunciations is suppressed.
\vspace*{-20pt}

\section{Experiments}
\label{sec:pagestyle}

\subsection{Dataset}
\label{ssec:subhead}

In our following experiments, the CMU \footnote{http://www.speech.cs.cmu.edu/cgi-bin/cmudict} pronunciation dictionary is used to evaluate the performance, which is a publicly authoritative dataset in G2P task. We split the dataset into train and test set as recommended in \cite{rao2015grapheme,Yolchuyeva2019,chen2003conditional} so that the results are directly comparable. After removing polysyllabic words and split, the CMU dataset contains a 106,837-word training set and a 12,000-word test set. 2,670 words are used as development (validation) set. There are 28 graphemes (lowercase alphabet symbols plus the apostrophe and hyphen) and 42 phonemes including $blank$ label used in CTC (84 phonemes considered stress and syllable boundary) in this dataset. The letter to phoneme mapping dictionary is designed by language expert.

\subsection{Software and hardware details}
\label{ssec:subhead}
One Nvidia V100 (32 GB) GPU card hosted in a 8 cores server with 32GB RAM served for training and inference phase. For training and evaluation, we use Pytorch \footnote{https://pytorch.org/} deep learning framework as our environment.

\subsection{Experimental configurations}
\label{ssec:subhead}

In LiteG2P, the two input features are firstly separately embedded into latent embeddings before feeding into the model and both of the embedding size is 64. For the 1D CNN in each CNN block, channel is 128, kernel size is 3 and stride is 1. For the bidirectional GRU block, we have experimented with three sets of models with different hidden sizes: 128, 192, 256 corresponding to LiteG2P-small, LiteG2P-medium and LiteG2P-large respectively. Moreover, we set dropout rate to 0.1 for each GRU layer. In training phase, adam \cite{DBLP:journals/corr/KingmaB14} is used as optimizer, the initial learning rate is set to 0.001 and is multiplied by 0.5 for every 5 epochs with total 50 epochs. The batch size is 128.


\begin{table}[!htb]
\caption{Performance and model size of LiteG2P}
\label{tab2}
\centering
\begin{tabular}{c|ccc}
\toprule
Model & small & medium & large \\
\midrule
WER$_S$(\%) & 35.4 & 34.1 & \textbf{33.6} \\
WER$_{SS}$(\%) & 36.4 & 35.1 & \textbf{34.3} \\
WER(\%) & 26.5 & 24.3 & \textbf{24.0} \\
\midrule
size & \textbf{0.6M} & 1.27M & 2.25M\\
\bottomrule
\end{tabular}
\end{table}


\vspace*{-15pt}
\begin{table}[!htb]
\caption{Time consuming of the different sizes of LiteG2P model on cloud devices.}
\label{tab3}
\centering
\begin{tabular}{c|cc}
\toprule
 & Xeon 8260 CPU(2.4GHz) & V100 GPU \\
\midrule
small model & \textbf{1.6ms/word} & \textbf{1.2ms/word} \\
medium model & 2.0ms/word & 1.5ms/word \\
large model & 3.1ms/word & 1.6ms/word \\
\bottomrule
\end{tabular}
\end{table}


\vspace*{-15pt}
\begin{table}[!htb]
\caption{Time consuming of LiteG2P-small model on mobile devices.}
\label{tab4}
\centering
\begin{tabular}{c|c}
\toprule 
Device type & Run time \\
\midrule
A53 (1.4GHz) & 6.38ms/word \\
A53 (2.0GHz) & 5.24ms/word \\
A73 (1.8GHz) & 3.30ms/word \\
A73 (2.2GHz) & 3.32ms/word \\
A78 (2.8GHz) & \textbf{0.71ms/word} \\
\bottomrule
\end{tabular}
\end{table}


\vspace*{-15pt}
\begin{table*}
\caption{Performance, size and speed comparison of different methods on the CMU dataset}
\label{tab5}
\centering
\begin{tabular}{c|cccc}
\toprule
Method & WER(\%) & size & run time(CPU) & run time(GPU) \\
\midrule
Encoder CNN, decoder Bi-LSTM (Model 5) \cite{yolchuyeva2019grapheme} & 25.13 & 14.5M & N/A & N/A \\
End-to-end CNN (Model 4) \cite{yolchuyeva2019grapheme} & 29.74 & 7.62M & N/A & N/A  \\
Encoder-decoder LSTM with attention (Model 1) \cite{yolchuyeva2019grapheme} & 28.44 & 12.7M & N/A & N/A \\
Transformer 3x3 \cite{Yolchuyeva2019} & \textbf{23.9} & 1.49M & 66ms/word & 2.8ms/word  \\
DBLSTM-CTC 128 Units \cite{rao2015grapheme} & 27.9 & 3M & 12ms/word & N/A  \\
DBLSTM-CTC 512 Units \cite{rao2015grapheme} & 25.8 & 11M & 64ms/word & N/A \\
\midrule
LiteG2P-medium & 24.3 & \textbf{1.27M} & \textbf{2.0ms/word}  & \textbf{1.5ms/word}  \\
\bottomrule
\end{tabular}
\end{table*}

\section{Results Discussion}
\label{sec:pagestyle}

Word Error Rate (WER) is used as the evaluation metric, which represents the percentage of words in which the predicted phoneme sequence does not exactly match the reference pronunciation and the number of word errors is divided by the total number of unique words in the reference. For the reason that stress information and syllable boundary is utilized in our TTS system, in following experiments, we also considered the WER$_S$ and WER$_{SS}$ evaluation metrics in addition to WER metric. WER$_S$ denotes the WER considering stress without syllable boundary and WER$_{SS}$ denotes the same but considering both stress and syllable boundary.

In table \ref{tab2}, we first compared the accuracy and model size of the three setups of LiteG2P model. The performance of WER$_S$ and WER$_{SS}$ tend to be worse than WER due to considering more complex output. We can see that the model performance gets better as the model size increases. The LiteG2P-large model performs best accuracy while LiteG2P-small model has the smallest size among the models. It is worth noting that all of the three models are very compact, and the small version of LiteG2P has only 0.6M parameters, which is very promising to be applied even to most extremely poor mobile devices.

After analyzing the model size and performance of LiteG2P, we further carried out experiments on running speed, the results are as shown in table \ref{tab3} and table \ref{tab4}. For running time evaluation, we randomly selected out 100 words from the dictionary and calculated the average time of the 100 words prediction time. We first conducted experiments on cloud devices: Xeon 8260 with 2.4GHz chip frequency as the CPU device and Nvidia V100 card as the GPU device. As shown in table \ref{tab3}, the running time increases as the model size gets larger on both CPU and GPU devices. However, even LiteG2P-large only costs 3.1ms to predict a word on CPU while 1.6ms on GPU, which is really quite fast. And we found that compared with GPU, the speed change of CPU is more obvious. We guess there may be two reasons. One is that we did not use batch prediction in this experiment so that we cannot take advantage of parallel computing of GPU. And another reason is that the storage interaction time is magnified due to the model computation being too fast. Given the exciting computing advantages on cloud devices, we further conducted experiments on mobile devices as shown in table \ref{tab4}. We chose LiteG2P-small as the mobile model and experimentalized on 5 kinds of chips which covers the low-to-high end chips, and we can see that even on the lowest chip, the small model only costs 6.38ms per word, which is really quite fast and promising to be applied in many offline scenarios.

We further conducted a comprehensive comparison with other state-of-the-art end-to-end methods as shown in table \ref{tab5}. Before analyzing the results, some notes should be declared: firstly, the data about speed is not presented in \cite{Yolchuyeva2019}, here we used an internal implemented model to test the speed. Secondly, for DBLSTM-CTC models in \cite{rao2015grapheme}, the speed data is not specified to be tested in which device, here it is tentatively considered to be on the CPU device as the upper limit of speed. And finally considering the trade-off of performance and speed, we chose LiteG2P-medium as the cloud model. We can see that compared with the best version of DBLSTM-CTC\cite{rao2015grapheme} models, LiteG2P exceeds 1.5\% in WER with 10 times smaller size and 30 times faster speed. And the accuracy is even comparable with the transformer model\cite{Yolchuyeva2019}, while the speed of LiteG2P is 33 times faster on CPU and over 1.8 times faster on GPU without any optimization, the size of LiteG2P is also smaller. LiteG2P shows the best performance and speed trade-off over all end-to-end methods, which we attribute to the effective fusion of data driven model learning and expert knowledge. We also made a simple case study of whether to introduce the expert dictionary. For the word \textit{cat}, the predicted phonemes are \textit{"K AE1 T F"} without the dictionary and the result is corrected to \textit{"K AE1 T"} with the dictionary, which shows the 
filter effect on impossible phonemes of expert knowledge.

\vspace*{-10pt}
\section{Conclusion}
\label{sec:typestyle}

We investigated a novel end-to-end G2P architecture named LiteG2P. Owing to additional expert knowledge introduction and the help of CTC loss function, the model has advantages of  high accuracy, parallelism, light weight, and fast computing. Thorough experiments on the CMU dataset have shown that LiteG2P is superior to the state-of-the-art CTC based method with 10 times
fewer parameters, and even comparable to the state-of-the-art transformer-based sequence-to-sequence model with less parameters and computation. The model also shows great potential for mobile applications. In the future, we will further explore the application of LiteG2P in multilingual direction and automatic generation of the expert dictionary.

\bibliographystyle{IEEEbib}
\bibliography{Template.bbl}

\end{document}